\documentclass[12pt,numbers]{elsarticle}

\usepackage{amssymb}
\usepackage{amsmath}
\DeclareMathOperator*{\argmax}{arg\,max}  

\journal{IFAC EAAI}

\usepackage{algorithm}
\usepackage{algpseudocode}
\usepackage{subcaption}
\usepackage{booktabs}
\usepackage{multirow}
\usepackage{amsmath, amsthm}
\usepackage{mdframed}
\theoremstyle{definition}

\newcommand{\Z}{\mathbb{Z}}
\newcommand{\mc}[1]{\mathcal{#1}}
\newcommand{\EE}{\mathbb{E}}
\newcommand{\NN}{\mathbb{N}}

\newcommand{\Scal}{\mathcal{S}}
\newcommand{\Acal}{\mathcal{A}}

\newenvironment{problemformulation}{
    \begin{mdframed}[linewidth=1pt] 
    \textbf{Problem Formulation (Wave Scheduling Problem)}\par
    \medskip
}{
    \end{mdframed}
}

\begin{document}

\begin{frontmatter}


\title{Optimizing Agricultural Order Fulfillment Systems: A Hybrid Tree Search Approach\tnoteref{label1}}
\tnotetext[label1]{This work was supported by Corteva Agriscience under the research collaboration agreement between Corteva Agriscience and the University of Illinois Urbana-Champaign.}

\author[1]{Pranay Thangeda}
\ead{pranayt2@illinois.edu}

\author[2]{Hoda Helmi}
\ead{hoda.helmi@corteva.com}

\author[1]{Melkior Ornik}
\ead{mornik@illinois.edu}

\affiliation[1]{organization={University of Illinois Urbana-Champaign},
           city={Urbana},
           postcode={61801}, 
           state={Illinois},
           country={USA}}

\affiliation[2]{organization={Corteva Agriscience},
           city={Indianapolis},
           postcode={46268}, 
           state={Indiana},
           country={USA}}

\title{} 

\begin{abstract}
  Efficient order fulfillment is vital in the agricultural industry, particularly due to the seasonal nature of seed supply chains. This paper addresses the challenge of optimizing seed orders fulfillment in a centralized warehouse where orders are processed in waves, taking into account the unpredictable arrival of seed stocks and strict order deadlines. We model the wave scheduling problem as a Markov decision process and propose an adaptive hybrid tree search algorithm that combines Monte Carlo tree search with domain-specific knowledge to efficiently navigate the complex, dynamic environment of seed distribution. By leveraging historical data and stochastic modeling, our method enables forecast-informed scheduling decisions that balance immediate requirements with long-term operational efficiency. The key idea is that we can augment Monte Carlo tree search algorithm with problem-specific side information that dynamically reduces the number of candidate actions at each decision step to handle the large state and action spaces that render traditional solution methods computationally intractable. Extensive simulations with realistic parameters—including a diverse range of products, a high volume of orders, and authentic seasonal durations—demonstrate that the proposed approach significantly outperforms existing industry standard methods.
\end{abstract}



\end{frontmatter}



\section{Introduction}
\label{sec:introduction}
The seed supply chain plays a pivotal role in global agricultural business, acting as the cornerstone for both plant breeding programs and agricultural sustainability. The importance of these seed stocks is underscored by the critical need for timely fulfillment of seed orders to meet specific planting windows, often mandated by the seasonal growth cycles of different crops. Failure to meet these strict timelines can lead to a host of downstream issues, including suboptimal crop yields and financial loss~\cite{chandrasekaran2014agribusiness}.

\begin{figure}[h!]
    \centering
    \includegraphics[scale = 0.45]{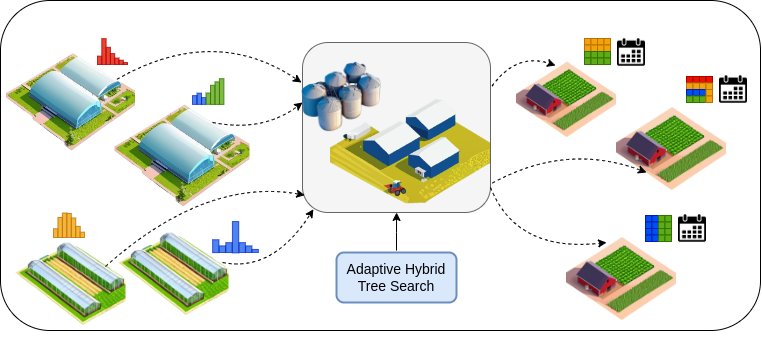}
    \caption{Overview of the centralized seed fulfillment process. The process begins with the arrival of seed stocks from multiple sites with stochastic, a priori unknown arrival distributions and ends with the fulfillment of orders with different deadlines and quantities. Our proposed adaptive hybrid tree search approach provides an efficient solution to the wave scheduling problem, optimizing the process of order fulfillment.}
    \label{fig:overview}
\end{figure}

Order fulfillment in industries such as e-commerce \cite{ricker1999order} and retail \cite{ishfaq2018evaluation} often involve centralized fulfillment centers that simultaneously process arriving inventory and fulfill orders based on their deadlines. The fulfillment process with large catalogs often handles a batch of orders, hereinafter referred to as \textit{wave}, together using automated sortation systems \cite{boysen2019automated}. The supply chain in these sectors is typically well-established, with known inventory quantities and deterministic restock times. The problem of optimally scheduling waves to maximize fulfillment efficiency is addressed using traditional operations research and optimization techniques \cite{liang2022estimation}, \cite{boysen2019warehousing} as order deadlines and inventory levels are known a priori or can be forecasted with low uncertainty. 

However, the seed distribution system presents a unique set of challenges that are not typically encountered in traditional supply chains. The nature of seed development and distribution involves a plethora of seed varieties, each with specific growing conditions and market demands. The perishability and seasonality of seeds make seed supply chains a highly intricate web of moving parts \cite{behzadi2018agribusiness}, \cite{yu2021disruption}. Recent advances in agricultural supply chain management \cite{jin2022impact, dong2023digitalization} have highlighted the growing complexity of seed distribution networks. Complicating matters further, the seed supply chain is inherently stochastic -- the variability in seed yields is influenced by numerous factors including weather conditions, soil quality, environmental factors \cite{behzadi2017robust}, crop diseases \cite{strange2005plant}, \cite{savary2019global}, and pest infestations \cite{suryawanshi2023distribution}, \cite{hardaker2015coping}.

Traditional optimization approaches \cite{moeller2011increasing}, \cite{zunic2017design}, \cite{van2018designing} often fall short in providing efficient solutions for seed wave scheduling with stochastic inventory arrivals. For instance, a simple greedy strategy that prioritizes fulfilling orders based on their deadlines and current available inventory can be misleading. Such an approach tends to focus on short-term gains, neglecting to consider the stochastic nature of future seed arrivals.

Markov Decision Processes (MDPs) \cite{puterman2014markov}, \cite{sutton2018reinforcement} serve as a useful framework to approach stochastic decision-making problems. MDPs have been widely applied in a range of domains such as robotics \cite{kober2013reinforcement}, \cite{zhao2020sim}, \cite{thangeda2022adaptive}, finance \cite{charpentier2021reinforcement}, \cite{hu2019deep}, transportation \cite{thangeda2020protrip}, \cite{zheng2021optimal}, healthcare \cite{yu2021reinforcement}, and large-scale infrastructure management \cite{thangeda2024infralib}, providing a systematic way to make optimal decisions in situations characterized by uncertainty. Practical considerations in MDP applications often involve resource constraints \cite{blahoudek2021fuel}, which in our context translates to inventory for order fulfillment. Recent work has explored deep reinforcement learning for warehouse scheduling \cite{mahmoudinazlou2025deep, ren2024dynamic}, but these approaches require extensive training data that is often unavailable in agricultural settings. However, directly applying traditional solution methods such as value iteration \cite{bellman1954theory} and policy iteration \cite{howard1960dynamic} to wave scheduling problem with large state and action spaces is computationally intractable due to the curse of dimensionality \cite{bellman1957markovian}. Instead, Monte Carlo tree search (MCTS) \cite{kocsis2006bandit} provides an intelligent tree search method that is effective in large MDPs where exhaustive search is not feasible \cite{browne2012survey}. Recent applications of MCTS in logistics \cite{swiechowski2023monte} demonstrate its effectiveness, but the exponential action spaces in wave scheduling remain challenging even for advanced MCTS variants.

To address this challenge, we propose an \textit{adaptive hybrid tree search approach} that augments MCTS with domain-specific knowledge to efficiently navigate the decision space. The key insight is that we can augment MCTS with additional problem-specific side information to make it tractable for large MDPs. The novelty of our approach lies not merely in using domain knowledge, but specifically in \textit{how} this knowledge is dynamically and adaptively integrated with the MCTS framework. This includes unique mechanisms such as the dual-criteria candidate order selection, the dynamic adjustment of the schedule based on MCTS feedback, and an adaptive action set generation process tailored for this domain. These elements go beyond generic heuristics, offering a structured and responsive way to manage the vast search space inherent in wave scheduling.

We validate our approach through extensive experiments, employing a high-fidelity warehouse simulator based on real-world data to showcase its efficacy. Our results indicate that the proposed method significantly outperforms traditional techniques, offering a robust, flexible, and efficient solution to the seed fulfillment scheduling problem.

The remainder of this paper is organized as follows: In Section \ref{sec:preliminaries}, we provide preliminaries and background on Markov chains and Markov decision processes. In Section \ref{sec:problem-formulation}, we formally define the problem statement and discuss the challenges of solving it. Section \ref{sec:solution-methodology} describes our approach to solving the problem using a combination of historical data, stochastic modeling, domain knowledge, and tree search methods. Finally, in Section \ref{sec:experiments-results} we present the experimental setup and discuss the results comparing our approach with traditional techniques.

\section{Preliminaries and Background}
\label{sec:preliminaries}

In this section we provide a brief overview of the mathematical structures that form the basis of our approach. We first introduce Markov chains, followed by time-varying Markov chains and Markov decision processes. We begin by defining the notation used in the paper.

Given a finite set \( \mathcal{A} \), we use \( |\mathcal{A}| \) to denote the cardinality of the set. $\Pr[\cdot]$ denotes the probability of an event and $\EE[\cdot]$ denotes the expectation of a random variable. $\NN$ denotes the set of natural numbers and $[N]$ for $N \in \NN$ denotes the set $\{0, \ldots, N -1 \}$. \( \mathbb{I}[.] \) denotes the indicator function, a function that takes a condition as its argument and returns 1 if the condition is true, and 0 otherwise.

\subsection{Markov Chains}

A Markov chain \cite{norris1998markov} is a stochastic model of transition dynamics defined over a sequence of states to represent systems that follow the \textit{Markov property}, i.e., systems in which the state at any given time depends solely on the state at the previous time step. Formally, a finite-state discrete-time Markov chain is defined by:

\begin{itemize}
    \item a finite set \(\mathcal{S} = \{s^1, s^2, \ldots, s^n\}\), representing all possible states the system can attain, and
    
    \item a square matrix \(\mathcal{P} \in [0,1]^{n \times n}\), where the entry \(\mathcal{P}_{ij}\) denotes the probability of transitioning from state \(s^i\) to \(s^j\).
\end{itemize}           

\subsection{Time-Varying Markov Chains}
Markov chains generally assume stationary transition probabilities, which may not always be suitable for modeling systems where the transition probabilities evolve over time \cite{bekaert1995time}, \cite{hosseini2012time}. Time-Varying Markov Chains (TVMCs) \cite{howard2012dynamic} are a generalization of Markov chains that allow the transition probabilities to vary over time. 

A TVMC retains the essential structure of a Markov chain but extends it by making the transition probabilities a function of time \(t\). Formally, instead of a static \(\mathcal{P}\), it introduces \(\mathcal{P}_{t}\), where \(\mathcal{P}_{t}(s^i, s^j)\) is the transition probability from state \(s^i\) to \(s^j\) at time \(t\), i.e., $\Pr(s_{t+1} = s' | s_t = s, t) = \mathcal{P}_{t}(s, s').$

\subsection{Markov Decision Processes}

Markov Decision Processes (MDPs)  \cite{puterman2014markov}, \cite{bellman1954theory} generalize Markov chains by introducing actions and rewards, thereby creating a framework for decision-making in stochastic environments. An \textit{undiscounted}, \textit{finite-horizon} MDP is formally defined as \( M(\mathcal{S}, \mathcal{A}, \mathcal{P}, \mathcal{R}, H) \), where:

\begin{itemize}
    \item \(\mathcal{S}\) denotes the finite state space analogous to that in Markov chains,
    
    \item \(\mathcal{A}\) denotes the finite action space, with $\Acal(s) \subset \Acal$ denoting the set of actions available in state $s \in \Scal$,
    
    \item ${\mathcal{P}: \Scal \times  \Acal \times \Scal \to [0,1]}$ satisfying $\sum_{s' \in \Scal} \mathcal{P}(s'|s,a) = 1$ for any $s \in \Scal$ and $a \in \Acal$ denotes the state transition probability function,
    
    \item $\mathcal{R}: \Scal \times \Acal \times \Scal \to [R_{min}, R_{max}]$ denotes the bounded reward function, quantifying the immediate reward for each transition,
    
    \item H is the finite planning horizon for the problem.
\end{itemize}

Taking an action $a \in \Acal(s)$ at a state $s \in \Scal$ results in a transition to a new state $s' \in \Scal$ with probability $\mathcal{P}(s'|s,a)$ and a reward $\mathcal{R}(s, a, s')$. The goal is to maximize the expected sum of rewards over planning horizon $H$.

A \textit{policy} \( \pi: \mathcal{S} \rightarrow \mathcal{A} \) is a mapping from states to actions such that $\pi(s) \in \Acal(s)$ for all $s \in \Scal$. The \textit{state value} function of a policy \( \pi \) for a given state \( s \) is defined as the expected sum of rewards over the next \( H \) time steps, i.e.,
\begin{equation}
V_H^\pi(s) = \mathbb{E} \left[ \sum_{k=0}^{H} \mathcal{R}(s_k, \pi(s_k), s_{k+1}) \bigg| s_0 = s \right].
\end{equation}
The \textit{state-action} value function, denoted by \( Q_H^\pi(s, a) \), is defined as the expected sum of rewards over the next \( H \) time steps, given that the first action is \( a \) and the first state is \( s \), i.e.,
\begin{equation}
Q_H^\pi(s, a) = \mathbb{E} \left[ \sum_{k=0}^{H} \mathcal{R}(s_k, \pi(s_k), s_{k+1}) \bigg| s_0 = s, \pi(s_0) = a \right].
\end{equation}

The optimal policy \( \pi^* \) is one that maximizes \( V_H^\pi(s) \) irrespective of the initial state, i.e., \(V_H^{\pi^*}(s) \geq V_H^\pi(s) \) for all \( s \in \mathcal{S} \). 

In the next section, we define the seed processing warehouse scheduling problem and formalize the problem statement.

\section{Problem Formulation}
\label{sec:problem-formulation}
\begin{figure}[t!]
    \centering
    \includegraphics[scale = 0.90]{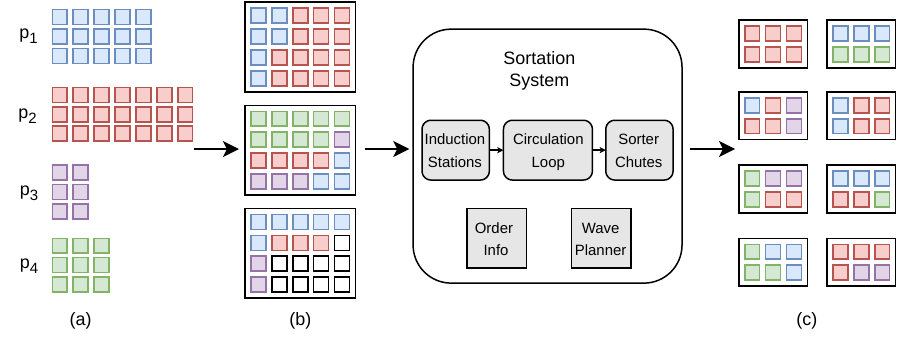}
    \caption{Illustration of the order fulfillment process in the warehouse with (a) incoming product quantities, (b) items placed in containers, (c) orders fulfilled by the system.}
    \label{fig:formulation-illus}
\end{figure}

We consider the setting of a warehouse processing incoming inventory of different products to fulfill orders made up of one or more products before fulfillment deadlines known a priori. The incoming inventory is arranged into \textit{intermediate containers}, which are fixed size storage containers. The warehouse utilizes an automated sorter system \cite{boysen2019automated} that can process a batch of orders, hereinafter referred to as \textit{wave}, using accumulation chutes temporarily assigned to individual orders. Processing a wave requires human pickers to pick the required items from intermediate containers at the sorter's induction stations and place them on the sorter's circulation loop. Figure \ref{fig:formulation-illus} provides an illustration of a simple fulfillment scenario where all the inventory required to fulfill the orders is available a priori.

Several existing approaches in the literature consider different objectives for optimally scheduling the orders in waves \cite{liang2022estimation}, \cite{boysen2019warehousing}, \cite{ii2000evaluation}, \cite{de2007design}. These approaches assume that the order information and deadlines are available along with the inventory required to fulfill these orders. This assumption, however, is unrealistic in our setting where the inventory arrives over the duration of entire season as the harvesting progresses with no prior knowledge of the incoming inventory's distribution \cite{kukal2018climate}. 

Formally, let $T$ represent the total number of discrete time steps in the planting season, each with a duration of $\Delta t$. Consider a set of products $P = \{p_1, p_2, \ldots, p_{n_p}\}$ in the catalog. We assume that the quantity of each product replenished in the inventory at any time step is stochastic and unknown a priori. As a consequence, the total incoming quantity of all products is stochastic. Let $q^p_t \in \Z$ denote the quantity of product $p$ that arrived at $t$.

Let $n_c$ denote the capacity of each intermediate container. Let $C_t$ denote the set of all intermediate containers arriving at time $t$. Let $c^p_{t, i}$ denote the quantity of product $p$ in intermediate container $i$ arriving at time $t$. We note that the total quantity of all products in an intermediate container is less than or equal to the capacity of the intermediate container, i.e., $\sum_{p \in P} c^p_{t, i} \leq n_c$. Also, the total number of items of a product $p$ in all the intermediate containers arriving at time $t$ is equal to the quantity of product $p$ arriving at time $t$, i.e., $\sum_{i \in C_t} c^p_{t, i} = q^p_t$. The number of intermediate containers arriving at time $t$, i.e., $|C_t|$, can be calculated as $|C_t| = \lceil \frac{\sum_{p \in P}q^p_t}{n_c} \rceil$.

Let $O$ denote the set of orders, $O = \{o_1, o_2, \ldots, o_{n_o}\}$, where each order $o_i$ is defined by the quantities of each product needed, $b_i^p$ for $p \in P$, and the deadline $d_i$. All orders in $O$ are of the same size $n_b$, i.e., $\sum_{p \in P} b_i^p = n_b$ for all $i \in [n_o]$.  Let $n_w$ denote the wave capacity, i.e., the number of orders processed in a single wave. Assuming all the items required to fulfill orders in a wave are available, the time taken to fulfill a wave is proportional to the number of intermediate containers accessed by the pickers to fulfill the orders in that wave. 

In order to handle situations where it is not feasible to satisfy all order deadlines, we introduce a pre-emptive processing parameter, $\alpha$, ranging from $0.5$ to $1$. This parameter, typically user-defined, represents a degree of operational flexibility afforded to the system. The parameter $\alpha$ indicates the required proportion of available items in an order relative to the total number of items for initiating processing. Importantly, this form of preemptive processing is reserved for the last feasible opportunity to process an order without missing its deadline, effectively influencing the definition of a "feasible" order under such critical, time-sensitive conditions. Its role is to allow the system to process orders that are not fully complete but are very close to their deadline, where waiting for full availability would guarantee a missed deadline, thus balancing fulfillment rates with inventory availability. For instance, if $\alpha$ is $0.5$, order processing is triggered only when at least half of the items are available and processing at this point is the only way to avoid missing the deadline. This approach ensures the timely, albeit partial, fulfillment of orders, aligning closely with their respective deadlines.

Let $f_i$ denote the time step when an order $o_i$ is fulfilled. Let $W$ denote the set of all waves in the season, and let $\mathbf{w} = (w_{1,1}, w_{2,1}, \ldots, w_{{n_o},1}, \ldots, w_{{n_o},|W|})$ denote the decision variables where $w_{i,j} = 1$ if order $o_i$ is processed in wave $j$ and $w_{i,j} = 0$ otherwise. Given these definitions, the problem can be formulated as follows:

\begin{problemformulation}
  Given a set of orders and incoming inventory in a warehouse setting, minimize the total delay across all orders, defined as
  \begin{equation}
    \min_{\mathbf{w}} \ \ D = \sum_{i=1}^{n_o} \max(0, f_{i}(\mathbf{w}, \alpha) - d_{i})
  \end{equation}
  while satisfying the constraints ensuring that the required inventory is available to fulfill the orders, complete and preempted, at the beginning of each wave, each order is assigned to exactly one wave, and the number of orders in a wave is less than or equal to the wave capacity.
\end{problemformulation}

In this formulation, for each order $o_i$, we calculate the delay as the difference between the time step when it was fulfilled $f_{i}$ and its deadline $d_{i}$. If an order is fulfilled before its deadline, the difference $f_{i} - d_{i}$ will be less than or equal to zero, and we consider its contribution to the total delay as zero. Only when $f_{i} - d_{i}$ is greater than zero (i.e., the order is fulfilled after its deadline), it contributes to the total delay. 

The stochastic nature of inventory arrivals render conventional operations research and optimization techniques inadequate for our problem. In the next section, we discuss our approach that provides an efficient, approximate solution to the problem using a combination of historical data, stochastic modeling, domain knowledge, and tree search methods.

\section{Solution Methodology}
\label{sec:solution-methodology}

This section describes our approach to solving the wave scheduling problem. We begin by discussing the data available to us and the challenges of modeling the stochasticity in product arrivals. We then describe our approach to modeling the problem as a Markov Decision Process (MDP) and discuss the challenges of solving such an MDP at scale. Finally, we propose a novel hybrid tree search approach to overcome these challenges and provide an efficient solution to the wave scheduling problem.

\subsection{Stochastic Modeling of Product Arrival Distributions}
\label{sec:arrival-markov-chain}
Consider the historical data depicting the quantity of each product arriving at various times across seasons. Let \(m_p\) denote the number of seasons with historical data for a given product \(p\). The quantity of product \(p\) arriving at time \(t\) in the \(i\)-th season can be denoted as \(h_{p,t}^i\) where \(i \in [1, m_p]\). We note that \(m_p\) can vary across products, particularly when they have been introduced to the market at different times.

Given the historical data, the total quantity for product \(p\) during season \(i\) is \( \sum_{t=1}^{T} h_{p,t}^i \). We normalize these quantities as $
\tilde{h}_{p,t}^i = \frac{h_{p,t}^i}{\sum_{t=1}^{T} h_{p,t}^i} \times 1000$. Subsequently, we round \(\tilde{h}_{p,t}^i\) to the nearest integer. This transformation standardizes historical quantities for each product $p$ into a uniform integer range of 0 to 1000. This range ensures that the problem is computationally feasible while maintaining a sufficient resolution to capture the details in incoming inventory distribution.

\subsubsection*{Time-Varying Markov Chain Model}
To model the product arrival dynamics in the current season, we use a time-varying Markov chain for the cumulative quantity of each product $p$. The state space \(\mathcal{S}\) consists of integers from 0 to 1000, \(\mathcal{S} = \{0, 1, 2, \ldots, 1000\}\). Transition probabilities for each product $p$ are dictated by \(\mathcal{P}_{p,t}\), a time-dependent transition probability matrix. We determine \(\mathcal{P}_{p,t}\) based on historical data.

Let \(E_{p,t}^j(s \rightarrow s')\) represent the transition from state \(s\) at time \(t-1\) to state \(s'\) at time \(t\) during the \(j\)-th season for product \(p\). Similarly, let \(E_{p,t}^j(s)\) be the event that the state at time \(t-1\) was \(s\). The transition probability is then given by $
\mathcal{P}_{p,t}(s, s') = \frac{\sum_{j=1}^{m_p} \#(E_{p,t}^j(s \rightarrow s'))}{\sum_{j=1}^{m_p} \#(E_{p,t}^j(s))}$ where \(\#(E_{p,t}^j(s \rightarrow s'))\) denotes the number of times the event \(E_{p,t}^j(s \rightarrow s')\) occurred and \(\#(E_{p,t}^j(s))\) denotes the number of times the event \(E_{p,t}^j(s)\) occurred.

\begin{figure}[h]
  \centering
  \begin{subfigure}[b]{0.48\textwidth}
    \centering
    \includegraphics[width=0.85\linewidth]{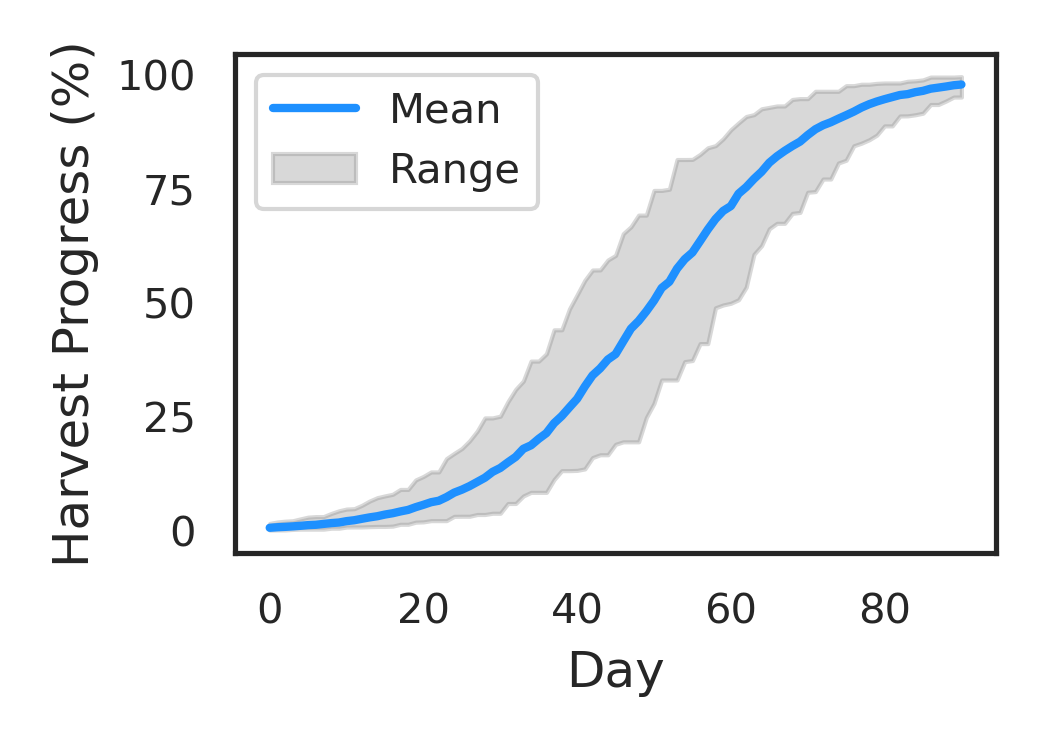}
    \caption{Historical harvest progress}
    \label{fig:harvest-progress}
  \end{subfigure}
  \begin{subfigure}[b]{0.48\textwidth}
    \centering
    \includegraphics[width=0.85\linewidth]{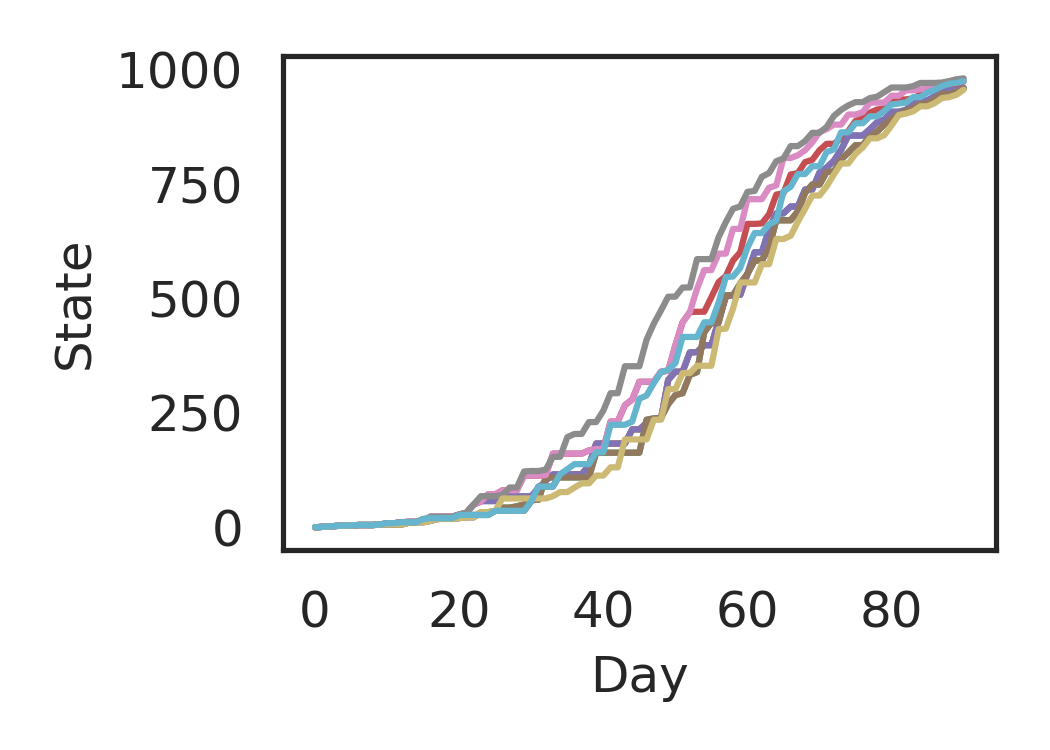}
    \caption{Simulated progress}
    \label{fig:harvest-sim}
  \end{subfigure}
 \caption{Illustration of the time-varying Markov chain prediction model estimated from historical data: (a) historical data depicting the harvest progress of corn in Iowa; (b) ten instances of simulated harvest progress using the model.}
  \label{fig:mc_illus}
\end{figure}

To illustrate the model's effectiveness in modeling crop harvest dynamics, we employ historical data on the harvest progress of corn in Iowa obtained from the United States Department of Agriculture (USDA) \cite{usda-nass2023}, as shown in Figure \ref{fig:harvest-progress}. This raw data was first normalized to our standardized state space $\mathcal{S}=\{0, 1, \ldots, 1000\}$, yielding $\tilde{h}_{p,t}^i$ for each past season. From these normalized historical trajectories, we empirically tallied the transition counts $\#(E_{p,t}^j(s \rightarrow s'))$ and $\#(E_{p,t}^j(s))$ across all seasons $j$ for each time step $t$. These aggregated counts directly inform the estimation of the time-varying transition probabilities $\mathcal{P}_{p,t}(s, s')$ using the formula provided. To generate the simulated harvest progressions in Figure~\ref{fig:harvest-sim}, each trajectory begins at state 0 (0\% harvested) at $t=0$. At each subsequent time step $t$, given the current simulated cumulative state $s_t$, the next state $s_{t+1}$ is sampled from the discrete probability distribution defined by the row $\mathcal{P}_{p,t}(s_t, \cdot)$. Repeating this iterative sampling process generates a complete simulated trajectory, and multiple such repetitions produce the ensemble of curves shown, demonstrating how the transition probabilities accurately capture and simulate to dynamic harvest trends.

\subsection{Wave Scheduling Problem as a Markov Decision Process}
We model the wave scheduling problem as a Markov Decision Process (MDP) to enable sequential decision-making while accounting for the product arrival uncertainty. Formally, the MDP model $M = (\mc{S}, \mc{A}, \mc{P}, \mc{R}, H)$ is defined as follows:

\subsubsection*{State Space}
To efficiently schedule the orders in waves, the state of the system which forms the basis for decisions must capture all the relevant information. The current inventory levels of all products dictate the feasibility of fulfilling orders in the planned wave. Furthermore, the time step and order deadlines are essential for both assessing and modeling the time-varying stochasticity of product arrivals. Accordingly, each state in the state space \( \mathcal{S} \), represented as \( s = (q^1, q^2, \ldots, q^{n_p}, d^1, d^2, \ldots, d^{n_p}, t) \), captures the current inventory levels of all products in the catalog, the deadlines, and the time step $t \in [T]$. 

\subsubsection*{Action Space}
At each decision step, the agent must decide which orders to process in the wave. The action space \( \mathcal{A} \), therefore, is the set of all possible waves of orders that can be processed. The set of available actions at state \( s \), \( \mathcal{A}_s \subseteq \mathcal{A} \), depends on \( s \) as the order feasibility is constrained by the available quantities of each product. Any action \( a \in \mathcal{A}_s \) represents a feasible wave of orders \( O_a \), i.e., $\sum \limits_{o_i \in O_a} b^p_i \leq q^p$ for any product $p$ that is a part of the orders \( O_a \). 

\begin{figure}[t!]
  \centering
  \includegraphics[scale = 0.75]{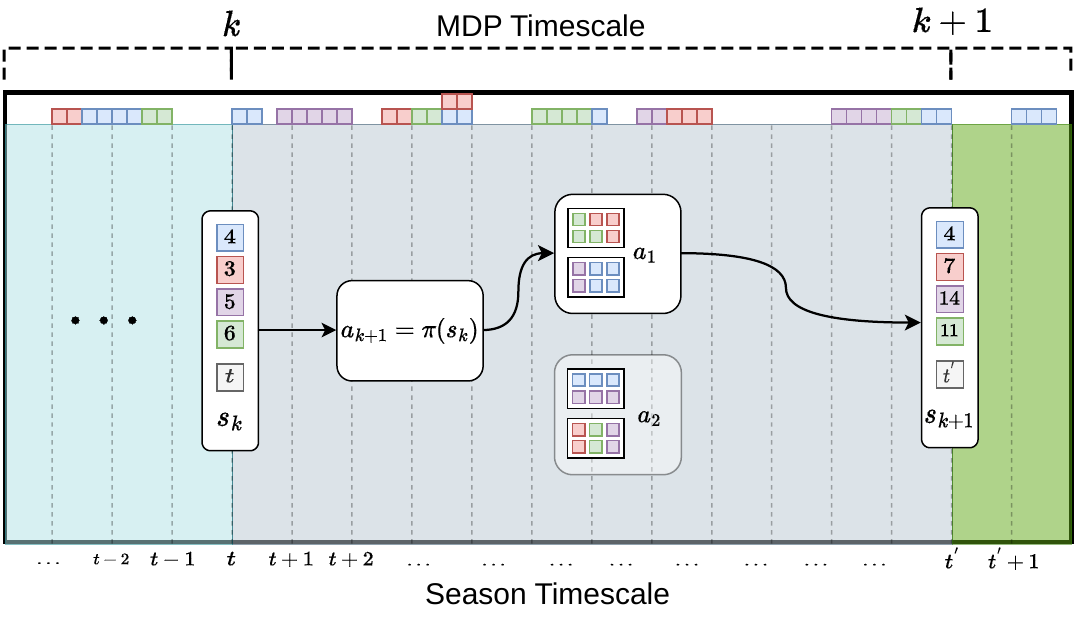}
  \caption{Illustration of the MDP formulation for the wave scheduling problem. The state space consists of the current inventory levels of all products and the time step. The action space consists of all feasible waves of orders. The transition probability function models the stochastic replenishment of inventory and the time taken to execute the wave.}
  \label{fig:mdp-illus}
\end{figure}

\subsubsection*{Transition Probability Function}
The transition probability function \( \mathcal{P} \) models how the system transitions from one state to another given an action. In the case of wave scheduling, after fulfilling the set of orders in a selected action, the system transitions to a new state based on the deterministic, known quantities of products consumed to fulfill the wave, the time taken to execute the wave, and the stochastic replenishment of inventory.

Formally, let $\mathcal{T}_a$ denote the random variable that represents the time taken to execute an action $a$ and let $\mathcal{Q}_{t,t'}^p$ denote the random variable that represents the stochastic replenished quantity of product $p$ between time steps $t$ and $t'$. For notational convenience, let $\mathcal{D}$ denote the set of all deadlines in the state space. Further, recall that the total quantity of product $p$ consumed by fulfilling orders $O_a$ specified by action $a$ is given by $\sum_{o_i \in O_a} b^p_i$. Then, given a state $s_k = (q^1_k, q^2_k, \ldots, q^{n_p}_k, \mathcal{D}, t_k)$ and an action $a_{k+1}$ at decision step $k$, the probability of transitioning to a new state \( s_{k+1} = (q^1_{k+1}, q^2_{k+1}, \ldots, q^{n_p}_{k+1}, \mathcal{D}, t_{k+1}) \) is given by
\[
\mathcal{P}(s_{k+1} | s_k, a_{k+1}) = \Pr(q^1_{k+1}, q^2_{k+1}, \ldots, q^{n_p}_{k+1}, \mathcal{D}, t_{k+1} | q^1_k, q^2_k, \ldots, q^{n_p}_k, \mathcal{D}, t_{k}, a_{k+1}),
\]
which can be further decomposed as follows:
\[
\mathcal{P}(s_{k+1} | s_k, a_{k+1}) = \Pr(\mathcal{T}_{a_{k+1}} = t_{k+1} - t_k) \times \prod_{p \in P} \Pr(\mathcal{Q}_{t_k, t_{k+1}}^p = q^p_{k+1} - q^p_k - \sum_{o_i \in O_a} b^p_i).
\]
The distribution of the quantity replenished for each product $\mathcal{Q}_{t_k, t_{k+1}}^p$ is determined by the time-varying Markov chain model discussed in Section \ref*{sec:arrival-markov-chain} and the distribution of $\mathcal{T}_{a_{k+1}}$ is established using empirical data specific to the configuration of the sortation system.

Note that we use \( t \) to indicate a time step in the discretized season duration, which is distinct from the decision step $k$ in the MDP. The decision step $k$ in the MDP is the instant at which the agent takes $k$-th action in the sequential decision-making process. Figure \ref{fig:mdp-illus} illustrates the MDP formulation for the wave scheduling problem while highlighting the distinction between the decision step and the time step.

\subsubsection*{Reward Function}
The objective of the wave scheduling problem is to ensure that all the orders are fulfilled before deadline, and in cases where this is not possible, to minimize the delay. We design the reward function $\mathcal{R}$ to reflect this objective. 

\( \mathcal{R}(s, a, s') \) specifies the immediate reward for taking action \( a \) in state \( s \) and transitioning to \( s' \). We calculate the reward as the number of orders met before their deadlines minus a penalty for the late orders proportional to the delay, weighted by a factor \( \lambda > 0 \):
\begin{equation}
\mathcal{R}(s, a, s') = \sum_{o_i \in O_a} \mathbb{I}[f_i \leq d_i] - \lambda \sum_{o_i \in O_a} \max(0, f_i - d_i)
\end{equation}

We now proceed to discuss the challenges of solving the above MDP formulation in practice. 

\subsubsection*{Challenges of Solving the Wave Scheduling MDP}

Existing methods for solving an MDP \( M \), given known transition probabilities \( \mathcal{P} \) and a reward function \( \mathcal{R} \), include dynamic programming-based approaches, deep learning-based techniques, and tree search methods that learn through interactions from a simulator of system dynamics. However, applying these algorithms to the wave scheduling MDP is challenging for the following reasons:

\paragraph{Large State and Action Spaces}
In real-world wave planning problems, the substantial size of the state and action spaces presents significant computational challenges. The state space \( \mathcal{S} \) in the wave scheduling MDP consists of all possible combinations of product quantities and the time step, with a cardinality of \( |\mathcal{S}| = \prod_{p \in P} (q^p + 1) \times T \). The action space at each state, \( \mathcal{A}_s \), includes all feasible waves based on the available product quantities, pending orders, and their deadlines. The size of the complete action space \( \mathcal{A} \) is \( |\mathcal{A}| = \sum_{s \in \mathcal{S}} |\mathcal{A}_s| \), which is exponential in the number of products and orders. 

For dynamic programming based approaches like value iteration, the computational complexity of each iteration is quadratic in the size of the state space and linear in the size of the action space which for the wave-scheduling problem translates to \( O\left(\prod_{p \in P} (q^p_{\text{max}} + 1) \times T \times \sum_{s \in \mathcal{S}} |\mathcal{A}_s|\right) \), rendering dynamic programming approaches intractable for our problem. Similarly, tree search methods like MCTS are also computationally infeasible due to the large branching factor of the tree associated with large action spaces. Although deep learning-based approaches offer greater scalability by generalizing performance across state and action spaces, they necessitate a compact representation of the state space and an extensive dataset encompassing the entire action space, which is impractical in our context.

\paragraph{Imperfect Transition Model}
The stochastic arrival of products is modeled using time-varying Markov chains, with associated transition probability matrix estimated from historical data, as discussed in Section \ref{sec:arrival-markov-chain}. However, the transition probability models are often susceptible to inaccuracies due to the lack of sufficient historical data, the presence of outliers, and the inherent stochasticity of the crop harvest dynamics. This imposes additional challenges on dynamic programming based approaches since each iteration might require recalculating the value function, exacerbating the already high computational cost.

In the next section, we propose an approach to overcome these challenges and solve the wave scheduling MDP in practice at real-world scale.

\subsection{Adaptive Hybrid Tree Search for Wave Scheduling}
To overcome the computational challenges inherent in solving the wave scheduling problem, we propose a hybrid approach that combines domain knowledge with tree search methods. The central idea of our approach is to combine the intelligent tree search and \textit{anytime} properties of general Monte Carlo Tree Search (MCTS) with domain-specific knowledge to dynamically curate a narrowed-down action space at each state to increase the likelihood of identifying promising actions under practical computational constraints.

In rest of this section, we delve into specifics of our approach, structured into three main components. Firstly, we present our approach to identifying and refining the feasible orders into a set of candidate orders. Next, we outline our method for constructing a practical and effective set of wave actions from the narrowed-down set of candidate orders. Finally, we elaborate on how we adapt MCTS algorithm in the context of our problem to parallely learn dynamic parameters that guide action set reduction process and tree search.

\subsubsection*{Candidate Order Set Selection}
At any decision step in an episode, let $O_{uf}$ denote the set of unfulfilled but feasible orders, i.e., orders that have not been fulfilled but can be fulfilled given the current inventory levels. It is important to note that an order might be in $O_{uf}$ at some point but still exceed its deadline if it is not selected for processing by the algorithm; this can occur due to strategic long-term decisions made by the MCTS, such as prioritizing other more critical orders or waiting for more items to form a larger, more efficient wave, based on its lookahead capabilities. In order to reduce the action space size, we generate two subsets $O_{uf}^{d}$ and $O_{uf}^{p}$ from $O_{uf}$. Orders for $O_{uf}^{d}$ are selected based on their deadlines in ascending order, giving precedence to the most urgent ones, while ensuring that orders already past their deadlines do not constitute more than a predetermined proportion of $O_{uf}^{d}$. For $O_{uf}^{p}$, orders are selected based on their proximity to the peak of unfulfilled order deadlines (the time step with the maximum number of currently unfulfilled orders), aiming to alleviate potential future bottlenecks. If an order meets the criteria for both subsets, it is primarily considered for $O_{uf}^d$ due to deadline urgency; the constraint $|O_{uf}^d \cap O_{uf}^p| = \emptyset$ ensures distinctness, typically by removing an order from consideration for $O_{uf}^p$ if it's already prioritized in $O_{uf}^d$. We constrain $|O_{uf}^d| \leq 2n_w$ and $|O_{uf}^p| \leq 2n_w$. The intuition is that $O_{uf}^d$ prioritizes orders with earlier deadlines, while $O_{uf}^p$ focuses on orders likely to cause bottlenecks due to deadlines clustering around the peak. This decomposition attempts to balance urgency of individual orders against long-term risk of developing system-wide bottlenecks.

To construct the candidate order set \(O_c\) of size $2n_{w}$, we introduce a dynamic parameter called the \textit{peak reducing factor} \(\rho \in [0,1]\). The value of $\rho$ varies based on tree search evaluation outcomes from previous rounds; its adaptation based on MCTS feedback is part of the learning process discussed in Algorithm \ref{alg:AdaptiveHybridTreeSearch}. $\rho$ adaptively adjusts the proportion of orders selected from $O_{uf}^d$ and $O_{uf}^p$ while ensuring a minimum number from each set. $O_c$ consists of the first $2n_w\rho$ orders from $O_{uf}^d$ and the first $2n_w(1-\rho)$ orders from $O_{uf}^p$. When \(|O_{uf}^d| < 2n_w\rho\), we add the remaining orders from $O_{uf}^p$ to $O_c$, and conversely, when \(|O_{uf}^p| < 2n_w(1-\rho)\), we add the remaining orders from $O_{uf}^d$ to $O_c$. This structured approach to forming $O_c$ by segmenting orders based on urgency and potential bottlenecks ($O_{uf}^d$ and $O_{uf}^p$) and then dynamically balancing their inclusion via $\rho$, is a key way domain-specific knowledge is integrated to prune the vast action space, making MCTS computationally tractable for this complex scheduling problem. Further refinement of the action space occurs during \textit{action set generation}, where waves are constructed from $O_c$ and filtered by estimated fulfillment time.

\subsubsection*{Action Set Generation}
While $O_c$ provides a reduced set of candidate orders, generating an appropriate and computationally tractable set of wave actions remains a challenge. Given that the size of $O_c$ is $2n_w$, the combinatorial possibilities yield \( \frac{2n_w!}{(n_w!)^2}\) candidate waves for the action space. The action set for our tree search must be manageable in size, only contain waves that are feasible at currently inventory levels, and filled with actions likely to be beneficial for maximizing our reward.

We adopt a two-step approach to generate the action set. First, we construct a set of feasible candidate waves by incrementally adding randomly selected orders from $O_c$ that ensure wave feasibility until the size of the set reaches wave capacity $n_w$. Next, we calculate the estimated fulfillment time of each candidate wave and select a predetermined number of waves with the lowest estimated fulfillment time. While this approach may not always yield the optimal set of candidate actions, it provides a computationally tractable way to generate the action set which is critical for our tree search algorithm that involves simulating multiple rollouts from each action.

\subsubsection*{Adaptive Tree Search}
In this section we discuss how we use the proposed adaptive hybrid tree search algorithm to solve the wave scheduling problem. Our approach closely follows the steps involved in Upper Confidence Bound for Trees (UCT) \cite{kocsis2006bandit} variant of MCTS with the modifications that we proposed in the previous sections. We begin with a brief overview of UCT algorithm and then proceed to discuss our approach to adapt it for the wave scheduling problem.

MCTS involves iteratively building a search tree to identify the best possible action to execute at the root node. It consists of four steps: (i) selection, (ii) expansion, (iii) simulation, and (iv) backpropagation. The tree starts as a single node, representing the current state. At each iteration, in the selection step, the algorithm starts at the root and moves down the tree according to a \textit{tree policy} until it reaches a leaf node. Then, if the leaf node is not a terminal state, the expansion step adds a child node to the tree, representing possible next state, by taking an action. The simulation step simulates a random rollout from a newly expanded node until it reaches a terminal state, resulting in a sequence of rewards. Finally, the backpropagation step propagates these reward values back up the tree, updating the state-action value estimate $\hat{Q}(s,a)$ and visit count of each node it traversed on the way. 

The UCT algorithm proposed the following tree policy to effectively balance between exploring new states and exploiting known, valuable states:
\begin{equation}
\pi_{tree}(s) = \argmax_{a \in A} \left[ \hat{Q}(s,a) + c \sqrt{\frac{\log \#(s)}{\#(s,a)}} \right],
\end{equation}
where $\hat{Q}(s,a)$ is the current estimate of state-action value, $\#(s)$ is the visit count of state $s$, $\#(s,a)$ is the visit count of state-action pair $(s,a)$, and $c$ is a constant that controls the amount of exploration.

In the context of the wave scheduling problem, at each decision step where we need to select an action, we initiate a tree with the root node representing the current state that encapsulates information about the current inventory levels of all products and the current time step. Then, we generate the action set on the go at each state and use the UCT tree policy to select the action to execute until we reach a leaf node and then expand the tree. We then simulate a rollout from the newly expanded node until we reach a terminal state, randomly selecting actions from generated reduced action sets at each state. Finally, we backpropagate the reward obtained from the rollout that depends on the number of orders that were fulfilled before their deadline and that missed their deadline. We perform iterations for a fixed time budget that depends on the wave completion time in the real system, and then execute the recommended action from the tree with the highest estimated state-action value at the root node in the warehouse. In order to execute the selected wave, we select the smallest set of available intermediate containers that contain all the products required to fulfill the orders in the wave and move them to the induction station of the sortation system. Note that this recommended action is the only action (wave) that is executed in the real sortation system. Once the wave is executed in the real system and transitions to the next state, we repeat the above steps before taking the next action. Algorithm \ref{alg:AdaptiveHybridTreeSearch} summarizes our approach to solving the wave scheduling MDP using UCT and selective action spaces.

\begin{algorithm}
  \caption{Adaptive Hybrid Tree Search for Wave Scheduling}
  \label{alg:AdaptiveHybridTreeSearch}
  \begin{algorithmic}[1]
  \State \textbf{Input:} Current system state $s$, list of orders $O$, wave capacity $n_w$, and peak reducing factor $\rho$
  
  \State Initialize tree with root node representing state $s$
  \State Initialize state-action rewards $Q$ and counts $\#$
  
  \State \textbf{Candidate Order Set Selection:}
  \State Identify unfulfilled and feasible orders $O_{uf}$ based on $s$
  \State Sort orders in $O_{uf}$ by deadlines and proximity to peak deadline to form $O_{uf}^d$ and $O_{uf}^p$
  \State Construct candidate order set $O_c$ using $\rho$ from $O_{uf}^d$ and $O_{uf}^p$
  
  \State \textbf{Action Set Generation:}
  \State Generate a set of potential wave actions $\mathcal{A}_s$ based on $O_c$ and $n_w$
  
  \State \textbf{Tree Search for Wave Scheduling:}
  \For{each iteration within the time budget}
      \State Select actions from $\mathcal{A}_s$ using $\pi_{tree}$ until leaf node
      \State Execute $a$ from $\pi_{tree}$ at leaf node to get new state $s'$
      \State Expand the tree by adding new node with state $s'$
      \State Perform a rollout simulation from $s'$ to a terminal state
      \State Calculate the rewards $r$ from the rollout
      \State Update $Q$ and $\#$ for along the path from root to leaf node
  \EndFor
  
  \State \textbf{Determine Best Action:}
  \State Choose the action $a^*$ as $a^* = \argmax_{a \in \mathcal{A}_s} Q(s,a)$
  
  \State \textbf{Output:} Action $a^*$ to be executed in the real system
  \end{algorithmic}
\end{algorithm}

The adaptive hybrid tree search approach provides a tractable way to solve the wave scheduling problem at scale while planning over a long horizon. The online planning nature of the approach also ensures that we can react online to the discrepancies between the estimated and actual transitions of the system.
Now that we have presented our approach to solving the wave scheduling problem, we proceed to discuss the experiment setting and compare the performance of the proposed approach with the baseline greedy approach currently used in practice.

\section{Experiments and Results}
\label{sec:experiments-results}

In this section, we present a comprehensive evaluation of our proposed approach against established baseline methods. We begin with a description of the experimental setup, followed by the baseline approaches used for comparison. We then present the performance results, conduct a sensitivity analysis on key parameters, analyze detailed scheduling performance by order category, and discuss the computational considerations of our method.

\subsection{Experimental Setup}
\label{sec:experimental-setup}

We conduct our experiments in a simulation environment designed to replicate the operational dynamics of a real-world centralized seed processing facility. The simulator accurately models the core components of warehouse operations including inventory management, order processing, and automated sortation, parameterized to align closely with real-world data to ensure relevance and applicability of our findings.

We consider a seed distribution season spanning 90 days, with a time step size of 1 minute to capture fine-grained information in the sortation system. As discussed in Section \ref{sec:problem-formulation}, the sortation system receives inventory in intermediate containers with a capacity of 250 items, and we assume the facility has sufficient capacity to accommodate all containers. The standard configuration features 30 induction stations and 400 chutes, which limits the maximum number of orders processed in a single wave to 400. An order is fulfilled when all its items reach the designated chute, and a wave concludes only when all its orders are fulfilled.

\begin{figure}[h!]
  \centering
  \includegraphics[width=0.9\columnwidth]{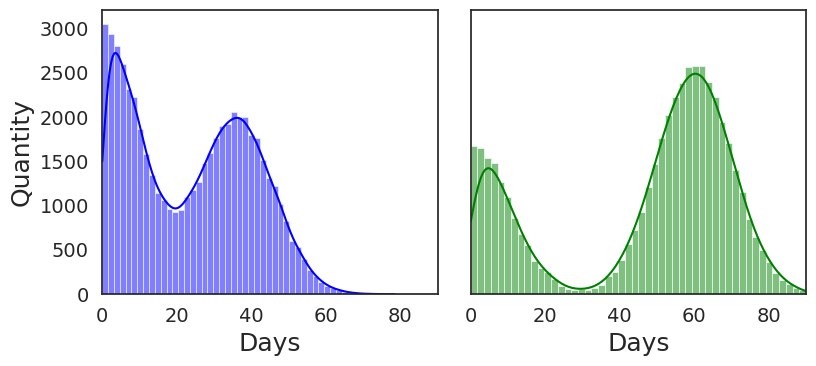}
  \caption{Illustrative example of the arrival distribution of two different products over the season. (left) The product has significant inventory available at the beginning of the season when compared to the production in the current season. (right) The product has significantly more inventory arriving in the current season when compared to the inventory available at the beginning of the season.}
  \label{fig:arrival-distribution}
\end{figure}

The simulation features 200 products in the catalog, each with a unique arrival distribution throughout the season. The quantity of each product arriving over the season is consistent with the quantity needed to fulfill all orders. We model the arrival distribution using a bimodal Gaussian distribution with peaks at the beginning and middle of the season, reflecting real-world scenarios where initial inventory comes from previous season's storage, while new production from the current harvest peaks mid-season. Figure \ref{fig:arrival-distribution} illustrates the arrival distribution of two representative products. Inventory arrivals are scheduled every 8 hours, consistent with typical operational practices.

\begin{figure}[h]
  \centering
  \includegraphics[width=0.96\columnwidth]{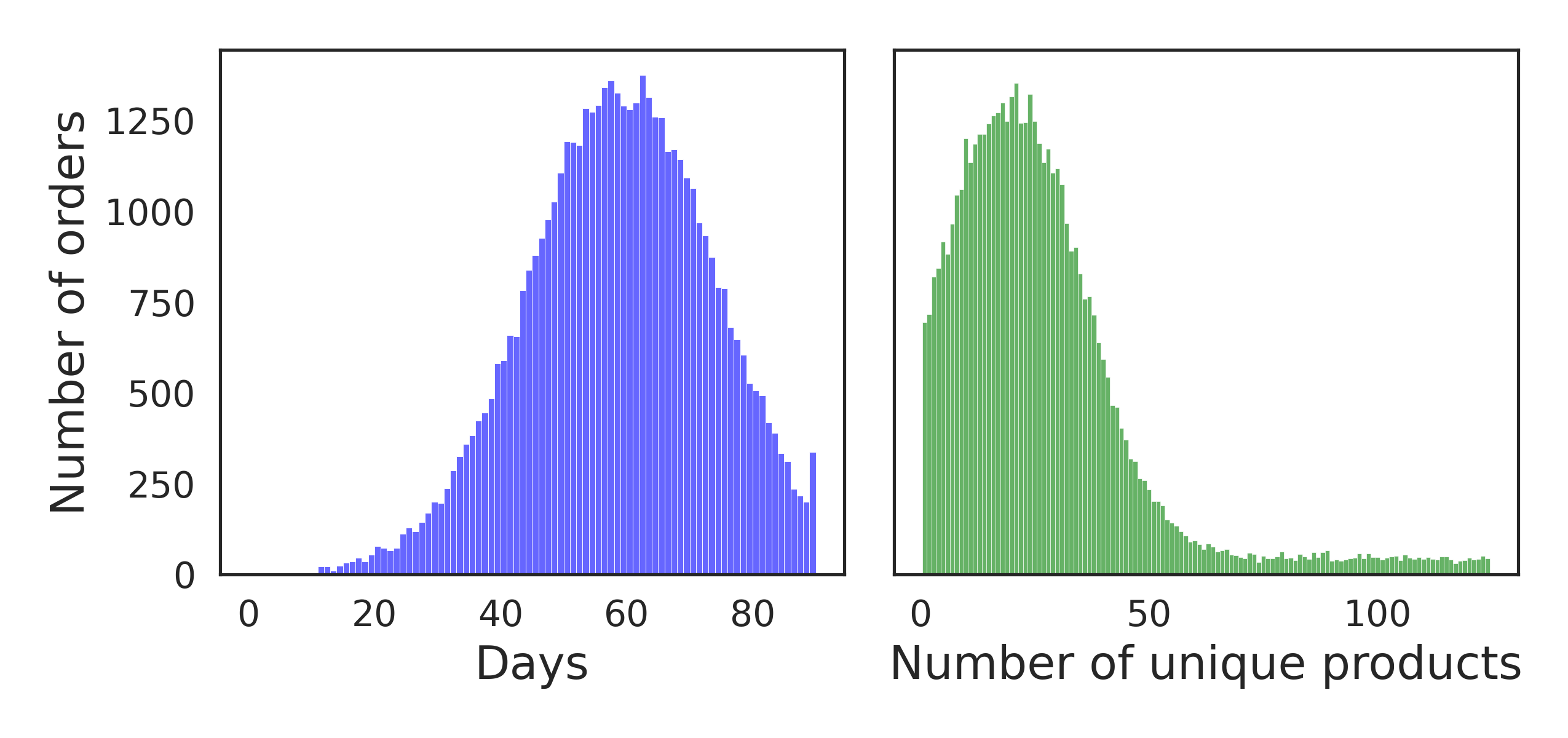}
  \caption{Illustrative example of order deadlines and compositions in a season.(left) Distribution of order deadlines over the season. (right) Distribution of the number of unique products in each order.}
  \label{fig:order-distributions}
\end{figure}

We simulate 50,000 orders where each order contains 250 items. For each order, a random set of products are selected with the number of unique products limited to a maximum of half the order size. We model the due dates of the orders to follow a Gaussian-like distribution with peak order volume in the middle of the season, mirroring real-world order patterns. Figure \ref{fig:order-distributions} shows the distribution of order deadlines and the number of unique products in each order in a typical season simulation run.

\subsection{Baseline Approaches}
We compare our proposed method against three baseline approaches:

\subsubsection*{Greedy Heuristic}
The first baseline is a greedy heuristic approach (GH) to allocate products to orders that is commonly used in real-world scheduling problems due to its simplicity and ease of implementation. This approach works as follows: first, the set of unfulfilled orders $O_{uf}$ are sorted by their due dates in ascending order. From this feasible set, we select the first $2n_w$ orders to form the candidate order set $O_c$. Then, the order with the earliest deadline in $O_c$ is selected and added to the current wave. Products are allocated to this order from inventory to fulfill while ensuring that the number of intermediate containers accessed is minimized. Next, within the remaining orders in $O_c$, the order that minimizes the number of intermediate containers accessed is chosen and added to the wave. This process is repeated, adding orders to the wave, until the wave is full or no more feasible orders remain. Once a wave is finalized, the wave is executed and next wave is constructed using the same steps with the latest inventory information. As a greedy method, this approach lacks foresight into future inventory arrivals and cannot take actions that are better in the long term.

\subsubsection*{Rolling Horizon Heuristic}
\label{sec:rolling-horizon-heuristic}
The Rolling Horizon Heuristic (RHH) represents a more sophisticated approach that incorporates lookahead capability into the scheduling process. At each decision point, RHH considers a fixed planning horizon that spans multiple future waves or a specified time window. Within this horizon, it attempts to optimize a sequence of waves through a multi-step process that begins with forecasting expected inventory arrivals based on historical arrival patterns. The heuristic then generates a set of candidate waves for the entire window, initially ordering them based on order deadlines.

The core of RHH's optimization process involves selecting waves across the limited horizon to minimize a comprehensive objective function. This function accounts for multiple factors: the predicted delays for orders included in these waves, penalties for not selecting near-deadline orders that could feasibly be processed within the window, and the estimated efficiency of container access for each potential wave configuration. After determining the optimal sequence of waves for the current horizon, RHH executes only the first wave from this sequence before advancing time and repeating the entire process with a new planning horizon.

While RHH incorporates a degree of planning and anticipation of future states that makes it more sophisticated than the purely reactive GH approach, it lacks the extensive stochastic lookahead and adaptive learning of our MCTS-based proposed approach.

\subsubsection*{Vanilla MCTS Baseline}
To validate the effectiveness of our domain-specific enhancements, we implement a vanilla MCTS baseline that uses standard UCT without our adaptive action space reduction mechanisms. Given the computational intractability of exploring the full action space, this baseline employs random sampling to maintain feasibility. Specifically, at each decision step, the vanilla MCTS randomly samples 1000 feasible waves, then applies the standard UCT algorithm to this randomly selected action set.

This implementation uses the same UCT formula, simulation budget, and computational resources as our proposed approach, with the key difference being the absence of intelligent action curation. Instead of our dual-criteria candidate order selection, dynamic peak reducing factor, and estimated fulfillment time filtering, the vanilla MCTS relies solely on random sampling to reduce the action space to a manageable size. This design choice allows us to isolate and quantify the impact of our domain-specific enhancements while maintaining computational tractability.

\subsection{Performance Results}
\label{sec:main-results}

This section presents the main experimental results comparing the performance of our proposed approach with the baseline methods. In order to account for the stochastic nature of the problem, we report statistics over 30 independent runs. Further, to ensure a fair comparison, we use the same set of order and product arrival distributions for both approaches in each unique run.

Given that our goal is to maximize the total number of orders fulfilled before their deadline, we use the percentage of orders fulfilled before their deadline without any preemptive processing ($\alpha = 1$) as the evaluation metric. We also report the delay statistics in orders that were fulfilled after their deadline.

\begin{table}[h]
  \centering
  \caption{Comparison of order fulfillment performance with $\alpha = 1$ calculated over 30 independent runs. Delay statistics are calculated for delayed orders only.}
  \label{tab:results_revised}
  \begin{tabular}{@{}lccc@{}}
    \toprule
    Approach & \% Orders Fulfilled & Avg. Delay & Median Delay \\
    {} & On Time ($\pm$ std) & (days, $\pm$ std) & (days, $\pm$ std) \\
    \midrule
    Greedy Heuristic (GH)     & 59.5\% ($\pm$ 2.5) & 9.6 ($\pm$ 1.2) & 8.5 ($\pm$ 1.0) \\
    Vanilla MCTS & 64.3\% ($\pm$ 2.3) & 8.7 ($\pm$ 1.1) & 7.9 ($\pm$ 0.9) \\
    Rolling Horizon (RHH) & 68.2\% ($\pm$ 2.2) & 7.5 ($\pm$ 1.0) & 6.8 ($\pm$ 0.8) \\
    Proposed Approach     & \textbf{81.7\%} ($\pm$ 1.8) & \textbf{4.2} ($\pm$ 0.8) & \textbf{3.5} ($\pm$ 0.6) \\
    \bottomrule
  \end{tabular}
\end{table}

Table \ref{tab:results_revised} summarizes the results from the experiment, clearly demonstrating the superiority of our proposed approach across all metrics. Our method achieved an 81.7\% on-time fulfillment rate, substantially higher than the 59.5\% achieved by GH, the 64.3\% achieved by vanilla MCTS, and the 68.2\% by RHH. The performance advantage extends to delay metrics as well, with our approach reducing the average delay for late orders to just 4.2 days, compared to 9.6 days with GH, 8.7 days with vanilla MCTS, and 7.5 days with RHH.

The vanilla MCTS results are particularly revealing. Despite using the same computational budget and UCT algorithm as our proposed approach, the vanilla MCTS achieves only marginal improvements over the simple greedy heuristic. This modest 4.8 percentage point improvement (from 59.5\% to 64.3\%) demonstrates that MCTS alone, without domain-specific guidance, struggles to effectively navigate the vast action space. The random sampling of waves, while computationally necessary, fails to consistently identify high-quality scheduling decisions, resulting in performance that falls well short of both RHH and our proposed method.

This significant performance improvement stems from the proposed approach's ability to effectively plan over longer horizons and anticipate future inventory arrivals. Unlike the baselines, our approach can strategically delay processing certain orders when it anticipates more efficient batch processing opportunities in the near future or when it detects potential bottlenecks ahead. For example, when faced with multiple approaching deadline clusters, our method can sacrifice processing less urgent orders to ensure capacity for an impending rush of near-deadline orders.

The performance gap between RHH and GH confirms the value of even limited lookahead planning. More importantly, the substantial 17.4 percentage point improvement of our approach over vanilla MCTS conclusively demonstrates that intelligent action space curation is essential for making MCTS effective in this domain.

\subsection{Sensitivity Analysis}
\label{sec:sensitivity-analysis}

To examine the robustness of our approach across different operational conditions, we conducted sensitivity analyses varying key parameters: the preemptive processing parameter ($\alpha$) and three warehouse constraints (container capacity, induction stations, and number of chutes/maximum wave size).

\begin{figure}[h!]
  \centering
  \includegraphics[width=\textwidth]{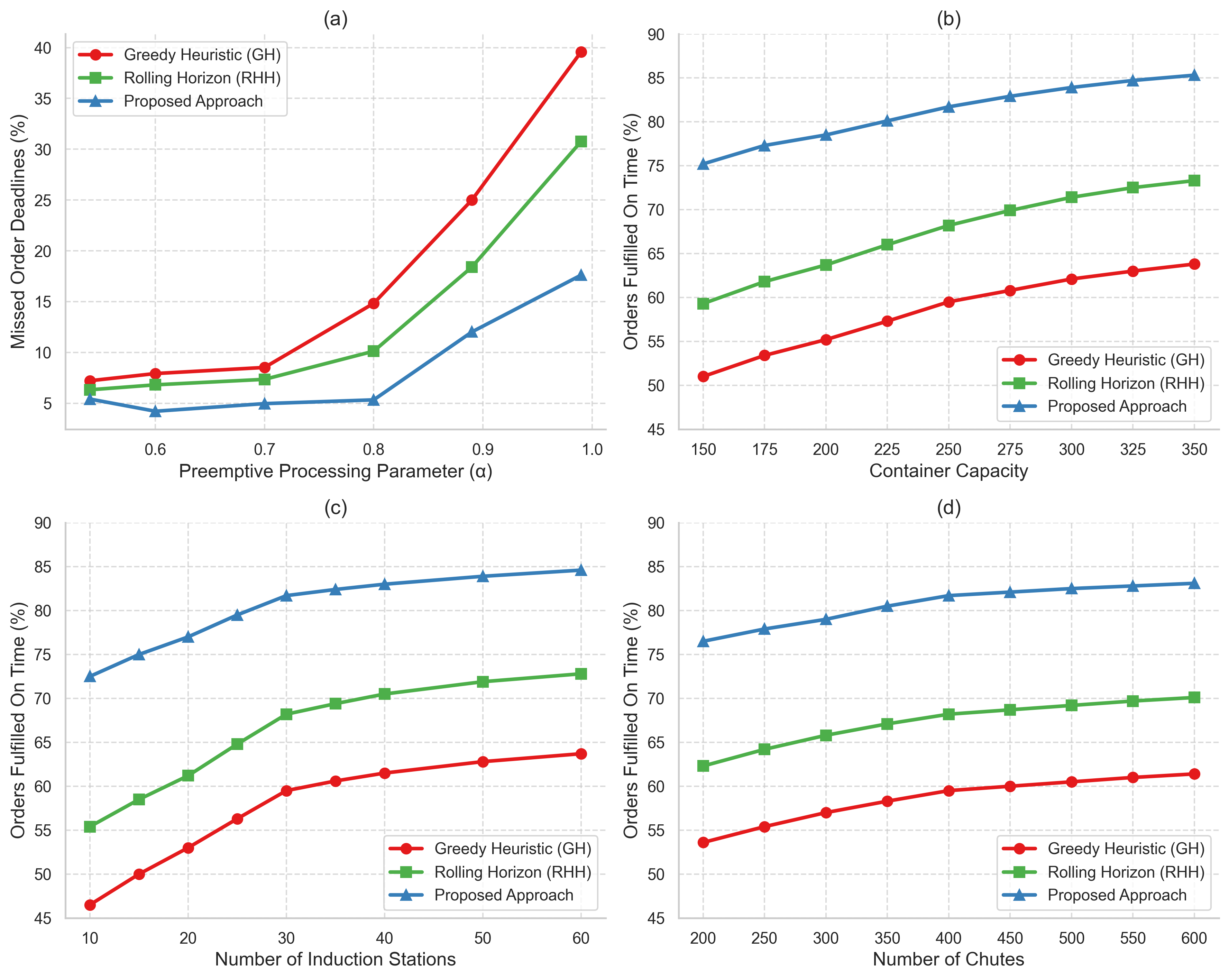}
  \caption{Sensitivity analysis of key parameters: (a) preemptive processing parameter ($\alpha$), (b) container capacity, (c) number of induction stations, and (d) number of chutes (maximum wave size). For the preemptive processing parameter, lower missed order rates indicate better performance. For the warehouse constraints, higher on-time fulfillment percentages indicate better performance. All metrics are visualized using line plots with distinctive markers for each approach (circle for GH, square for RHH, triangle for proposed method). Consistent coloring is used across all panels: red for Greedy Heuristic, green for Rolling Horizon Heuristic, and blue for our proposed approach.}
  \label{fig:sensitivity_analysis}
\end{figure}

\subsubsection*{{Effect of Preemptive Processing Parameter}}
\label{sec:alpha-sensitivity}

Figure \ref{fig:sensitivity_analysis}a shows the percentage of missed order deadlines across different values of $\alpha$ for all three approaches. Lower values of $\alpha$ allow processing orders with partial inventory when the deadline is approaching, while higher values require more complete inventory availability before processing.

The results demonstrate that our proposed approach consistently outperforms both baseline methods across all values of $\alpha$, with the performance gap widening as $\alpha$ increases. When $\alpha$ approaches 0.5, the performance gap narrows because the relaxed order completion requirements simplify the scheduling task. However, as $\alpha$ increases toward 1.0 (requiring full order completion), the scheduling problem becomes more challenging, and the sophisticated planning of our approach shows greater benefits. Statistical tests confirm that the performance differences are significant at all tested $\alpha$ values.

\subsubsection*{Effect of Warehouse Constraints}
\label{sec:warehouse-constraints-sensitivity}

Figures \ref{fig:sensitivity_analysis}b, \ref{fig:sensitivity_analysis}c, and \ref{fig:sensitivity_analysis}d present the impact of varying warehouse physical constraints on order fulfillment performance across extended parameter ranges. These constraints directly affect the operational capacity and efficiency of the fulfillment system.

\paragraph{Container Capacity} Figure \ref{fig:sensitivity_analysis}b shows that increasing container capacity from 150 to 350 items consistently improves performance across all approaches, as larger containers reduce the total number of containers needed and enhance picking efficiency. Our proposed approach shows consistent superiority, maintaining approximately a 20 percentage point advantage in on-time fulfillment over GH and a 13 point advantage over RHH across all capacity settings. The performance gains show a diminishing returns pattern at higher capacities, suggesting that beyond a certain threshold, other system constraints become limiting factors.

\paragraph{Induction Stations} Figure \ref{fig:sensitivity_analysis}c illustrates how the number of induction stations affects how quickly items can be introduced into the sortation system. While increasing the number of stations improves performance for all methods, our approach demonstrates robust performance even under constrained conditions, achieving 72.5\% on-time fulfillment with only 10 stations compared to GH's 46.5\% and RHH's 55.4\%.

\paragraph{Number of Chutes} Figure \ref{fig:sensitivity_analysis}d shows how the number of chutes, which directly affects the maximum wave size ($n_w$), impacts performance across a range from 200 to 600 chutes. Increasing this parameter yields moderate performance improvements across all approaches, with our method maintaining its significant advantage throughout the range. This indicates that while larger waves can improve efficiency to some extent, intelligent wave formation and scheduling are more critical factors in performance.

Overall, these sensitivity analyses demonstrate that our approach's advantage is robust across various operational configurations and parameter ranges. Even under the most constrained conditions tested, the long-horizon planning and adaptive nature of our method maintain a substantial performance edge over both baseline approaches. The consistent pattern across all parameter variations highlights the fundamental superiority of our approach's planning capabilities rather than its reliance on specific operational conditions.

\subsection{Scheduling Performance by Order Urgency}
\label{sec:detailed-scheduling-results}

In order to provide more granular insight into how scheduling decisions impact different types of orders, we analyzed fulfillment statistics for orders categorized by their initial deadline proximity when first considered by the scheduler. This analysis helps understand how each approach handles trade-offs between urgent and less time-sensitive orders.
\begin{table}[ht]
  \centering
  \caption{Order fulfillment performance by initial deadline proximity (30 runs, $\alpha=1$). Performance shown as percentage of orders fulfilled on time and average delay in days.}
  \label{tab:detailed_scheduling}
  \begin{tabular}{@{}l@{\hspace{0.3em}}c@{\hspace{0.3em}}c@{\hspace{0.3em}}c@{\hspace{0.3em}}c@{}}
    \toprule
    \multirow{2}{*}{} Deadline & Greedy Heuristic & Vanilla MCTS & Rolling Horizon & Proposed \\
    \midrule
    Urgent      & 45.2\% / 5.1d & 52.1\% / 4.3d & 58.4\% / 3.5d & 75.8\% / 2.2d \\
    Medium & 60.5\% / 9.2d & 65.3\% / 8.5d & 69.1\% / 7.0d & 82.3\% / 4.1d \\
    Low  & 72.8\% / 12.5d & 75.5\% / 11.2d & 78.1\% / 9.8d & 87.0\% / 6.3d \\
    \bottomrule
  \end{tabular}
\end{table}

Table \ref{tab:detailed_scheduling} presents performance metrics across three deadline categories: Urgent (deadline $<$ 7 days), Medium Priority (deadline 7-21 days), and Low Priority (deadline $>$ 21 days). All approaches demonstrate better on-time performance for less urgent orders due to increased scheduling flexibility, with our proposed approach achieving superior results across all categories.

The most significant performance difference appears in urgent order handling, where our approach achieves 75.8\% on-time fulfillment compared to 45.2\% for GH and 58.4\% for RHH. This demonstrates our method's effectiveness in prioritizing time-sensitive orders. The average delays are also consistently lower with our approach - for urgent orders, we reduce delays to 2.2 days versus 5.1 and 3.5 days for GH and RHH respectively.

Our approach shows more balanced performance across order categories, with only a 11.2 percentage point difference between low-priority and urgent orders. While GH's simple deadline-based approach struggles with urgent orders when resources are misaligned, and RHH shows limited improvement with its short planning horizon, our proposed method excels by balancing priorities between urgent and less critical orders, achieving superior performance across all categories.

\subsection{Computational Considerations}
\label{sec:computational-considerations}

Real-time warehouse applications necessitate careful consideration of computational demands for the proposed approach. The MCTS algorithm's complexity per decision step is approximately $N \times D_{avg} \times C_s$, where $N$ is the simulation count, $D_{avg}$ is their average depth, and $C_s$ is the simulation step cost. Our domain-specific heuristics for candidate order selection (sorting, $O(|O_{uf}|\log |O_{uf}|)$) and action generation (sampling and filtering) are designed for efficiency, managing $C_s$ and reducing the MCTS branching factor.

\begin{table}[ht]
  \centering
  \caption{Average computation time per decision step with standard deviations.}
  \label{tab:comp_time}
  \begin{tabular}{@{}lc@{}}
    \toprule
    Algorithm & Avg. Time per Decision (sec) \\
    \midrule
    Greedy Heuristic (GH)         & 0.42 ± 0.08 \\
    Vanilla MCTS         & 10.85 ± 2.11 \\
    Rolling Horizon Heuristic (RHH) & 3.15 ± 0.85 \\
    Proposed Approach (MCTS)        & 11.37 ± 2.24 \\
    \bottomrule
  \end{tabular}
\end{table}

Table \ref{tab:comp_time} presents empirical computation times for a simple warehouse scenario. The Greedy Heuristic (GH) exhibits very fast and consistent performance (0.42 ± 0.08 sec), the Rolling Horizon Heuristic (RHH) shows moderate variability (3.15 ± 0.85 sec), while our approach requires longer but still practical computation time (11.37 ± 2.24 sec), configurable by adjusting the MCTS simulation budget.

Despite this higher computational cost of our approach, practical factors support its viability. Wave scheduling decisions typically occur at larger time scales, and MCTS's anytime property permits flexible interruption. Crucially, the substantial performance gains (e.g., $>20\%$ improvement in on-time fulfillment) often justify this cost, as late deliveries are expensive. Empirical tests confirm tractability in high-throughput settings with adequate inter-wave planning time.

\section{Conclusion and Future Work}
\label{sec:conclusion}

In this paper, we introduced an adaptive hybrid tree search algorithm to address the stochastic wave scheduling problem, with a particular focus on order fulfillment in the agricultural domain. The method blends Monte Carlo tree search with domain-specific knowledge to provide a computationally tractable solution to our problem with large state and action spaces. Our approach outperforms the baseline greedy approach in terms of order fulfillment efficiency, demonstrating its effectiveness in managing complex agricultural sortation systems. The flexibility of our approach extends beyond seed order fulfillment, making it a versatile solution that can be adapted to a wide range of problems in order fulfillment and other domains that involve sequential decision-making under uncertainty.

While the presented results are promising, it is important 
to acknowledge certain limitations of the current study. Our validation relies on simulation, and while our simulator closely mirrors real operations, testing in an actual agricultural fulfillment center remains necessary to verify practical effectiveness. Additionally, the product arrival model's accuracy depends heavily on historical data quality, making our approach vulnerable to unforeseen shifts in arrival patterns.

Building on this work, several avenues for future research warrant exploration. One key direction is enhancing real-time adaptability; this includes developing an online learning mechanism for the peak reducing factor $\rho$ using a Bayesian update scheme, enabling within-season adaptation to improve performance during atypical arrival patterns. Additionally, hierarchical action abstraction through order clustering could significantly reduce computational complexity and enable longer planning horizons. Finally, improving the robustness of the product arrival model to non-stationarity is important; this could involve researching techniques to make the TVMC more quickly adaptable to shifts in seasonal patterns not present in historical data, or integrating online learning mechanisms.


\section*{Acknowledgments}
The authors would like to thank Venkat Nemani for his valuable feedback and discussions that contributed to the development of this work.

\bibliographystyle{unsrt} 
\bibliography{references}

\end{document}